\documentclass{article}

 \usepackage[preprint]{neurips_2026}

\usepackage[utf8]{inputenc} % allow utf-8 input
\usepackage[T1]{fontenc}    % use 8-bit T1 fonts
\usepackage{hyperref}       % hyperlinks
\usepackage{url}            % simple URL typesetting
\usepackage{booktabs}       % professional-quality tables
\usepackage{amsfonts}       % blackboard math symbols
\usepackage{nicefrac}       % compact symbols for 1/2, etc.
\usepackage{microtype}      % microtypography
\usepackage{xcolor}         % colors
\usepackage{graphicx}

\begin{document}
% -------------------------------------------------------
% NepKANUN: A RAG-Based Nepali Legal Assistant
% \title{NepKANUN: Democratizing Legal Knowledge in Nepal through Retrieval-Augmented Generation}
\title{NepKANUN: A RAG-Based Nepali Legal Assistant}

\author{%
  Bhabuk Thapa \\
  Department of Artificial Intelligence \\
  Kathmandu University \\
  Dhulikhel, Nepal \\
  \texttt{vhabukthapa@gmail.com} \\
  \And
  Prasiddha Koirala \\
  Department of Artificial Intelligence \\
  Kathmandu University \\
  Dhulikhel, Nepal \\
  \texttt{prasiddhaf23@gmail.com} \\
  \And
  Ranjit Raut \\
  Department of Artificial Intelligence \\
  Kathmandu University \\
  Dhulikhel, Nepal \\
  \texttt{rautranjit916@gmail.com} \\
  \AND
  Sunil Regmi \\
  Department of Artificial Intelligence \\
  Kathmandu University \\
  Dhulikhel, Nepal \\
  \texttt{sunil.regmi@ku.edu.np} \\
  \And
  Bal Krishna Bal \\
  Department of Computer Science and Engineering \\
  Kathmandu University \\
  Dhulikhel, Nepal \\
  \texttt{bal@ku.edu.np} \\
}

% \author{%
%   Bhabuk Thapa \\
%   Department of Artificial Intelligence \\
%   Kathmandu University \\
%   Dhulikhel, Nepal \\
%   \texttt{vhabukthapa@gmail.com} \\
%   \And
%   Prasiddha Koirala \\
%   Department of Artificial Intelligence \\
%   Kathmandu University \\
%   Dhulikhel, Nepal \\
%   \texttt{prasiddhaf23@gmail.com} \\
%   \And
%   Ranjit Raut \\
%   Department of Artificial Intelligence \\
%   Kathmandu University \\
%   Dhulikhel, Nepal \\
%   \texttt{rautranjit916@gmail.com} \\
%   \And
%   Sunil Regmi \\
%   Department of Artificial Intelligence \\
%   Kathmandu University \\
%   Dhulikhel, Nepal \\
%   \texttt{sunil.regmi@ku.edu.np} \\
%   \And
%   Bal Krishna Bal \\
%   Department of Computer Science and Engineering \\
%   Kathmandu University \\
%   Dhulikhel, Nepal \\
%   \texttt{bal@ku.edu.np} \\
% }

% \author{
%   Ranjit Raut\inst{1} \and
%   Bhabuk Thapa\inst{1} \and
%   Prasiddha Koirala\inst{1}  \and
%   Sunil Regmi\inst{1}  \and
%   Bal Krishna Bal\inst{2}
% }

% \institute{
%   Department of Artificial Intelligence, Kathmandu University, Dhulikhel, Nepal \and
%   Department of Computer Science and Engineering, Kathmandu University, Dhulikhel, Nepal
% }

% \authorrunning{Raut et al.}
\pagestyle{plain}
\maketitle

% -------------------------------------------------------
% Abstract
% -------------------------------------------------------
\begin{abstract}
Accessing legal information in Nepal is difficult due to complex terminology, limited resources, and misinformation. We introduce an AI-powered legal assistant that is tailored for Nepali legal texts and is built on a fine-tuned large language model. The technology provides precise, streamlined answers to natural language legal inquiries when integrated into a Retrieval-Augmented Generation (RAG) framework. It was trained using a custom dataset of high-quality question-answer pairs, and according to BERTScore, it obtained strong F1 scores of 0.82 (simple), 0.77 (moderate), and 0.71 (complex). Its usability is further confirmed by expert reviews. Our method shows how merging generation and retrieval can effectively democratize access to legal knowledge in Nepal by focusing on customized legal data and incorporating RAG.

% \keywords{Large Language Model \and Retrieval-Augmented Generation \and Parameter-Efficient Fine-Tuning \and Natural Language Processing \and Low-Rank Adaptation \and Optical Character Recognition}
\end{abstract}

% -------------------------------------------------------
% Introduction
% -------------------------------------------------------
\section{Introduction}
\label{sec:introduction}
The intersection of Artificial Intelligence (AI) and law presents transformative opportunities to enhance access to justice and legal knowledge globally. However, people speaking low-resource languages like Nepali have suffered greatly as a result of these developments, which have disproportionately benefited high-resource language regions. The complexity of legal terminology and structures, along with low public knowledge and limited access to judicial resources, frequently makes it difficult for people in Nepal to grasp their legal rights and duties.

Although NLP has advanced information retrieval and question answering in various domains, its application to Nepali legal texts is constrained by the scarcity of annotated datasets and the specialized nature of legal language. The urgent need for customized AI solutions is highlighted by the fact that Nepal’s current legal information systems frequently rely on keyword matching and lack the semantic depth necessary to correctly comprehend user inquiries.

Large Language Models (LLMs), such as the Llama series, GPT-4, and others, have greatly improved natural language processing (NLP), allowing for more complicated task-solving capabilities and more natural human-computer interaction. Despite their ability to understand and generate text, even in languages like Nepali, LLMs may display factual errors or "hallucinations," particularly in fields that require a lot of expertise or are changing quickly. To address these problems, Retrieval-Augmented Generation (RAG) has surfaced, which combines LLMs with an external retrieval step and grounds responses in reliable sources prior to generation. This procedure enables the integration of domain-specific knowledge and establishes the output in validated facts, greatly enhancing accuracy and traceability, particularly for knowledge-intensive tasks.

Addressing the challenges of accessing legal information in Nepal, this paper’s key contributions are:
\begin{itemize}
    \item Creation of a domain-specific, superior dataset that includes Nepali legal question-answer pairs in order to fill the gap in annotated legal data.
    \item Design and implementation of a RAG-based system that generates precise, context-aware answers to legal queries by utilizing an LLaMA 3.2 3B model that has been refined on a unique Nepali legal QA dataset.
\end{itemize}

% -------------------------------------------------------
% Methodology
% -------------------------------------------------------
\section{Methodology}
\label{sec:methodology}

\subsection{Data Preparation}
A high-quality, domain-specific dataset was necessary to modify the LLM to answer Nepali legal questions. We began curating a dataset with over 16,000 entries gathered using a hybrid strategy that involved online scraping from the Supreme Court website and pertinent news sources as well as processing PDF legal documents. To align with common user queries, these texts were transformed into question-answer pairs specifically designed for instruction tuning using prompt engineering. PyTesseract for Devanagari script was used to digitize the scanned documents, and they underwent essential manual validation. A systematic data cleaning pipeline, involving deduplication and correction of OCR-induced errors, yielded the final 10,000 high-quality entries utilized for model fine-tuning.

\begin{table}[htbp]
\centering
\caption{Summary of the NepKANUN dataset used for fine-tuning. The dataset includes question-answer pairs curated from Nepali legal texts.}
\label{tab:corpus}
\small
\begin{tabular}{lcp{4cm}p{2.5cm}}
\toprule
\textbf{Source}            & \textbf{Entries} & \textbf{Type}          & \textbf{Validation} \\
\midrule
Supreme Court Website      & 8,000            & Q\&A Pairs             & Manual \\
Legal News Sources         & 4,000            & Q\&A Pairs             & Manual \\
PDF Legal Documents        & 4,000            & Raw Text              & OCR + Manual \\
\midrule
\textbf{Total}             & \textbf{16,000}  & --                     & -- \\
\midrule
\textbf{Final Dataset}     & \textbf{10,000}  & High-Quality Q\&A Pairs & Manual \\
\bottomrule
\end{tabular}
\end{table}

\subsection{Fine-Tuning}
We selected the Llama 3.2 3B instruct model, noting its balance of model size, performance characteristics, and suitability for multilingual instruction-following tasks. We employed Parameter-Efficient Fine-Tuning (PEFT), specifically Low-Rank Adaptation (LoRA) and Quantized LoRA (QLoRA). Low-rank matrices are introduced by LoRA to decrease trainable parameters, and the base model is quantized to 4-bit precision by QLoRA to further increase efficiency. The Unsloth library was used to implement these strategies, which were tuned for quicker LoRA fine-tuning and lower memory consumption.

% \begin{table}[htbp]
% \centering
% \caption{Hyperparameters for fine-tuning LLaMA 3.2 3B.}
% \label{tab:hyperparameters}
% \scriptsize
% \begin{tabular}{|l|l|}
% \hline
% \textbf{Parameter}         & \textbf{Value}         \\
% \hline
% Learning Rate              & 3e-4 and 2e-4          \\
% \hline
% Weight Decay               & 0.01                   \\
% \hline
% Scheduler                 & Cosine                 \\
% \hline
% Optimizer                 & AdamW (8-bit)          \\
% \hline
% Mixed Precision            & BFloat16               \\
% \hline
% Max Sequence Length        & 2048 tokens            \\
% \hline
% \end{tabular}
% \end{table}
\begin{table}[htbp]
\centering
\caption{Hyperparameters for fine-tuning LLaMA 3.2 3B.}
\label{tab:hyperparameters}
\begin{tabular}{ll}
\toprule
\textbf{Parameter}         & \textbf{Value}         \\
\midrule
Learning Rate              & 3e-4, 2e-4             \\
Weight Decay               & 0.01                   \\
Scheduler                 & Cosine                 \\
Optimizer                 & AdamW (8-bit)          \\
Mixed Precision            & BFloat16               \\
Max Sequence Length        & 2048 tokens            \\
\bottomrule
\end{tabular}
\end{table}

\subsection{RAG Methodology}
We used a Retrieval-Augmented Generation (RAG) architecture to make sure the produced responses are correctly based on reputable legal materials. RAG systems are especially well-suited for fields like law, where having access to accurate, validated data from a particular corpus is crucial.

Important Nepali legal texts, such as the Constitution of Nepal 2072 and portions of important legislation (such as the Environmental Act and Muluki Ain), served as the main source of information for the RAG system. PyTesseract for OCR was used to handle a large number of documents that were available as scanned PDFs. For efficiency, a caching mechanism was included. We used a structure-aware chunking technique that divided text according to these logical divisions in order to maintain the semantic structure present in legal documents, which are frequently arranged hierarchically (e.g., into Parts, Chapters, and Articles).

A multilingual SentenceTransformer model, based on architectures like Sentence-BERT, was used to transform processed text chunks into dense vector embeddings. This model was selected because it can capture semantic linkages across languages, including Nepali. A vector embedding of 384 dimensions was used to represent each piece. ChromaDB, an open-source vector database designed for effective similarity search, was used to index and store these embeddings.

Initially, a user's query is embedded into the same 384-dimensional vector space. The retrieval component does a similarity search to find the most relevant ChromaDB pieces. To enhance contextual coverage, we used Maximal Marginal Relevance (MMR) to choose the top \( k = 9 \) document chunks, weighing variety across the retrieved sections against relevance to the query. The generating component, the fine-tuned Llama 3.2 3B model, receives the retrieved document chunks concatenated with the initial user query. The LLM incorporates data from the given context to provide the final response. The model's prompts and queries were carefully developed in Nepali to guarantee useful and contextually relevant results.

\begin{figure}[htbp]
\centering
\includegraphics[width=0.7\textwidth]{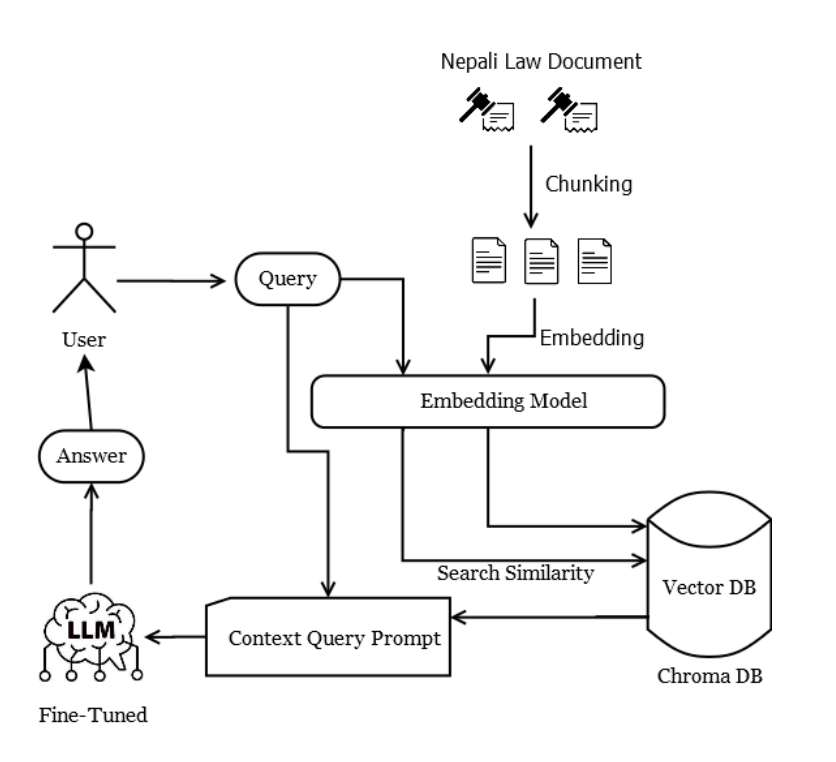}
\caption{Overview of the workflow for AI assistance in Nepali law.}
\label{fig:rag_workflow}
\end{figure}

% -------------------------------------------------------
% Results and Discussion
% -------------------------------------------------------
\section{Results and Discussion}
\label{sec:results}

\subsection{Automated Evaluation}
Objective metrics were used to assess output quality. While ROUGE and BLEU are common, their reliance on surface-level token overlap limits their effectiveness for evaluating semantic correctness, particularly in legal settings where paraphrasing can convey the same meaning. Therefore, we used BERTScore, which measures semantic similarity using contextual embeddings from pre-trained BERT models.

BERTScore provides Precision, Recall, and F1 values by comparing token embeddings, offering a more robust evaluation of semantic alignment. After evaluating the system across a spectrum of questions of varying complexity, we obtained the F1 scores shown in Table~\ref{tab:f1_scores}.

% \begin{table}[htbp]
% \centering
% \caption{F1 scores for different query complexities.}
% \label{tab:f1_scores}
% \scriptsize
% \begin{tabular}{|l|c|c|c|}
% \hline
% \textbf{Query Type}        & \textbf{Simple} & \textbf{Moderate} & \textbf{Complex} \\
% \hline
% F1 Score                   & 0.82            & 0.77              & 0.71              \\
% \hline
% \end{tabular}
% \end{table}

\begin{table}[htbp]
\centering
\caption{F1 scores for different query complexities.}
\label{tab:f1_scores}
\begin{tabular}{lccc}
\toprule
\textbf{Query Type} & \textbf{Simple} & \textbf{Moderate} & \textbf{Complex} \\
\midrule
F1 Score            & 0.82            & 0.77              & 0.71              \\
\bottomrule
\end{tabular}
\end{table}

\subsection{Human Evaluation}
A panel of Nepali lawyers and law students performed a human review in addition to the computerized one to evaluate the system's practicality. Five-point ratings were assigned to responses based on five important metrics: Faithfulness, Relevance, Logical Correctness, Completeness, and Interpretability.

% \begin{table}[htbp]
% \centering
% \caption{Evaluation ratings for a sample query across various metrics.}
% \label{tab:human_evaluation}
% \scriptsize
% \begin{tabular}{|l|c|}
% \hline
% \textbf{Metric}            & \textbf{Rating (out of 5)} \\
% \hline
% Faithfulness               & 4.5                       \\
% \hline
% Relevance                  & 5.0                       \\
% \hline
% Logical Correctness        & 4.0                       \\
% \hline
% Completeness               & 4.0                       \\
% \hline
% Interpretability           & 4.0                       \\
% \hline
% \end{tabular}
% \end{table}

\begin{table}[htbp]
\centering
\caption{Evaluation ratings for a sample query across various metrics.}
\label{tab:human_evaluation}
\small
\begin{tabular}{lc}
\toprule
\textbf{Metric}            & \textbf{Rating (out of 5)} \\
\midrule
Faithfulness               & 4.5 \\
Relevance                  & 5.0 \\
Logical Correctness        & 4.0 \\
Completeness               & 4.0 \\
Interpretability           & 4.0 \\
\bottomrule
\end{tabular}
\end{table}

\subsection{Discussion}
The outcomes demonstrate how well the algorithm handles simple legal questions. Strong performance in providing precise and query-aligned answers for factual questions is indicated by the BERTScore of 0.82 for basic queries and a flawless Relevance score of 5.0/5 in human evaluation. The RAG framework's ability to anchor responses in legal texts and guarantee factual consistency is further demonstrated by the high Faithfulness score of 4.5/5. However, when query complexity increases, a noticeable drop in performance is observed. For complicated queries, the BERTScore falls to 0.71, while the human ratings for Completeness and Logical Correctness decrease to 4.0/5. These findings suggest that the system struggles to synthesize and explain nuanced or ambiguous legal concepts, particularly in complex queries. This highlights the problem that previous legal NLP studies have shown, which is that automated measures frequently fall short of capturing complex legal thinking, particularly for ambiguous laws. Although the system is a useful tool for retrieving basic legal information, it has to be improved in order to handle more in-depth legal analysis.

% -------------------------------------------------------
% Conclusion
% -------------------------------------------------------
\section{Conclusion}
\label{sec:conclusion}
This work represents a significant advancement in legal AI for low-resource languages, as demonstrated through our Nepali legal assistant system. We have created a tool that provides precise and straightforward responses to legal questions in Nepali, based on reliable sources, by combining a RAG framework with a refined LLM. This work democratizes legal information by offering a workable way to improve legal accessibility in Nepal. Our system gives people a basic grasp of their legal rights and duties, but it does not take the place of expert legal assistance, particularly in complicated instances.

% -------------------------------------------------------
% Future Work
% -------------------------------------------------------
\section{Future Work}
\label{sec:future_work}
Future enhancements will focus on increasing the system's robustness and broadening its impact. Key areas include implementing mechanisms for automatic updates to the legal knowledge base to incorporate new laws and rulings, potentially through direct integration with official governmental databases, subject to privacy and technical constraints. Incorporating multimodal inputs like speech queries and extending language support to regional Nepali languages are also important directions aimed at creating a more accessible AI-driven platform for legal information in Nepal and potentially other low-resource contexts.

% -------------------------------------------------------
% Limitations
% -------------------------------------------------------
\section{Limitations}
\label{sec:limitations}
The effectiveness of RAG retrieval may be limited by the absence of embedding models that are especially trained for the Nepali legal area. Although the 10,000-pair Q/A dataset is of good quality, its size may limit generalization by failing to completely capture rare or extremely complicated legal issues. Peak model performance was probably affected by efficiency-focused fine-tuning constraints (QLoRA, trained on only two epochs) as opposed to more thorough training. Reliability in intricate legal reasoning may also be impacted by the model's difficulties with ambiguous or interpretable legal texts.

% -------------------------------------------------------
% Acknowledgments
% -------------------------------------------------------
\section*{Acknowledgments}
We would like to express our sincere gratitude to students of Kathmandu University School of Law for helping us by performing human assessment. 
% We would also like to thank the reviewers for their feedback and comments.

% -------------------------------------------------------
% Disclosure of Interests
% -------------------------------------------------------
\section*{Disclosure of Interests}
The authors have no competing interests to declare that are relevant to the content of this article.

% -------------------------------------------------------
% Bibliography
% -------------------------------------------------------
% \bibliographystyle{splncs04}
\bibliographystyle{unsrt}

\end{document}